\documentclass[letterpaper, 10 pt, conference]{ieeeconf}  

\IEEEoverridecommandlockouts                              

\usepackage{algorithm}
\usepackage{algorithmic}
\usepackage{subcaption}
\usepackage{amsmath} 
\usepackage{amsfonts} 
\usepackage{amsfonts}       

\usepackage[percent]{overpic}

\usepackage{multirow}
\usepackage{graphicx}
\usepackage{varioref}
\usepackage{tabularx}

\usepackage{booktabs}
\usepackage{dsfont}
\usepackage{xcolor}         
\usepackage[hyphens]{url}

\newcommand{\ie}{\textit{i.e.}}
\newcommand{\eg}{\textit{e.g.}}
\newcommand{\ours}{PCDT}

\title{\LARGE \bf
Predictive Coding for Decision Transformer
}

\author{Tung M. Luu$^\dagger$, Donghoon Lee$^\dagger$, and Chang D. Yoo$^*$ 
    \thanks{$^*$Corresponding author: Chang D. Yoo} %
    \thanks{$\dagger$ The authors are equally contributed.} %
    \thanks{All authors are with the School of Electrical Engineering, KAIST (Korea Advanced Institute of Science and Technology), Daejeon, Republic of Korea. \{tungluu2203,dh\_lee99,cd\_yoo\}@kaist.ac.kr}%
    \thanks{This work was partly supported by Institute for Information \& communications Technology Planning \& Evaluation (IITP) grant funded by the Korea government(MSIT) (No. 2021-0-01381, Development of Causal AI through Video Understanding and Reinforcement Learning, and Its Applications to Real Environments) and partly supported by Institute of Information \& communications Technology Planning \& Evaluation (IITP) grant funded by the Korea government(MSIT) [RS-2021-II212068, Artificial Intelligence Innovation Hub (Seoul National University)].}
}

\begin{document}

\maketitle
\thispagestyle{empty}
\pagestyle{empty}

\begin{abstract}

Recent work in offline reinforcement learning (RL) has demonstrated the effectiveness of formulating decision-making as return-conditioned supervised learning. Notably, the decision transformer (DT) architecture has shown promise across various domains. However, despite its initial success, DTs have underperformed on several challenging datasets in goal-conditioned RL. This limitation stems from the inefficiency of return conditioning for guiding policy learning, particularly in unstructured and suboptimal datasets, resulting in DTs failing to effectively learn temporal compositionality. Moreover, this problem might be further exacerbated in long-horizon sparse-reward tasks. To address this challenge, we propose the Predictive Coding for Decision Transformer (PCDT) framework, which leverages generalized future conditioning to enhance DT methods. PCDT utilizes an architecture that extends the DT framework, conditioned on predictive codings, enabling decision-making based on both past and future factors, thereby improving generalization. Through extensive experiments on eight datasets from the AntMaze and FrankaKitchen environments, our proposed method achieves performance on par with or surpassing existing popular value-based and transformer-based methods in offline goal-conditioned RL. Furthermore, we also evaluate our method on a goal-reaching task with a physical robot. 

\end{abstract}

\section{INTRODUCTION}

Reinforcement learning (RL) systems have demonstrated remarkable success across a wide array of domains, ranging from games \cite{silver2016mastering} to autonomous driving \cite{you2019advanced} and robotics \cite{kalashnikov2018qt,luu2021hindsight}. Yet, the inherent sample inefficiency of online RL algorithms presents a formidable challenge, demanding extensive environmental interactions for agent training. This necessity for voluminous data imposes substantial requirements for human supervision, safety checks, and resets \cite{atkeson2015no}, thereby limiting their practical deployment in real-world scenarios. Consequently, offline RL has recently garnered increasing attention as an alternative training paradigm, wherein policies are exclusively trained from static datasets of reward-labeled demonstrations. Offline RL algorithms commonly integrate learning objectives that encourage pessimism alongside value-based methods \cite{fujimoto2019off,kumar2020conservative,liu2020provably} to achieve commendable performance. Despite their promise, these methods encounter challenges during training, often necessitating meticulous hyperparameter tuning and various implementation tricks to ensure stability and optimal performance across various tasks.

In an effort to streamline offline RL training, recent research has explored transforming offline RL into a sequence modeling problem \cite{chen2021decision, janner2021sequence}, employing the powerful Transformer architecture \cite{vaswani2017attention} for decision-making. At the core of these transformer-based approaches lies the concept of conditioning policies on a desired outcome. For instance, Decision Transformer (DT) \cite{chen2021decision} learns a model to predict actions based on historical context, encompassing states and actions, and conditioned on a target future return. Through conditioning on future returns, DT effectively conducts credit assignment across time horizons, demonstrating competitive performance across various offline RL tasks. Importantly, these approaches eliminate the necessity for temporal-difference learning methods, such as fitted value or action-value functions. This results in a simpler algorithmic framework that relies on supervised learning, facilitating the advancement of offline RL by leveraging existing progress in supervised learning. 

Despite their potential, relying on reward-labeled datasets to train return-conditioned policies makes scaling DT challenging for large-scale training. Indeed, task-specific reward functions often require careful engineering for design \cite{ng1999policy, popov2017data}, making them costly to access in practice. Additionally, in sparse reward environments where rewards are often constant (\ie, success or failure), DT struggles to learn tasks, especially in long-horizon tasks \cite{correia2023hierarchical}. Furthermore, DT also faces difficulty in unstructured datasets that require the agent to learn temporal compositionality (\ie, stitching) by combining sub-trajectories of different demonstrations \cite{badrinath2024waypoint,kumar2022should}. For instance, in the AntMaze maze navigation environment, where an 8-DoF quadruped robot is required to reach the target, DT typically performs worse than other value-based offline RL methods such as IQL \cite{kostrikov2021offline}.

In this study, we delve into the offline goal-conditioned RL setting \cite{ding2019goal, ghosh2021learning, emmons2021rvs}. This setting allows the agent to solve multiple tasks by learning from reward-free data. We adopt the DT as an expressive policy network to facilitate goal-conditioned RL. However, the aforementioned challenges of DT in this setting still remain. To tackle these challenges, we introduce a framework that enables DT to condition on predictive codings for action predictions instead of returns. Specifically, we first learn a goal-conditioned representation that encodes the target goal, past, and future trajectory information using a bidirectional transformer. Next, we employ a causal transformer that takes as input the sequence of state-action pairs and conditions on the predictive codings from the previous stage to produce actions. The predictive codings serve as behavioral guidance toward the desired goal during policy learning. Our approach bypasses the requirement of return sequences, allowing the use of easy-to-collect offline data devoid of rewards for greater efficiency. By conditioning on these predictive codings, our transformer-based policy is equipped with the ability to reason about the future, providing crucial guidance for learning in unstructured and suboptimal datasets, leading to competitive performance across goal-conditioning benchmarks, particularly for long-horizon tasks. To summarize, our main contributions include:
\begin{enumerate}
    \item Introducing Predictive Coding for Decision Transformer (\ours), a framework enabling transformer-based agents to learn from a large set of diverse, unstructured, and suboptimal demonstrations by conditioning on predictive codings, thereby bypassing the requirement of rewards from the dataset.
    \item Proposing an effective method to learn predictive codings from datasets without requiring rewards and actions. Our results demonstrate that predictive coding effectively guides transformer-based agents to solve goal-conditioned tasks without the need for returns and enables stitching capability for DT.
    \item Evaluating \ours \ on eight datasets from two complex long-horizon goal-conditioning environments, AntMaze and FrankaKitchen, and validating it on the physical 7-DOF Sawyer robot. The code can be found at \url{https://github.com/tunglm2203/pcdt}.
\end{enumerate}

\section{RELATED WORK}

Offline reinforcement learning (RL) aims to learn effective policies from pre-collected datasets. Previous works have addressed the distribution shift problem \cite{fujimoto2019off} of offline RL through various strategies such as constraining the policy action space \cite{fujimoto2019off,kumar2019stabilizing}, incorporating value pessimism \cite{kumar2020conservative,liu2020provably}, or leveraging dynamics models \cite{kidambi2020morel,yu2020mopo}. In this paper, we focus on offline RL methods based on conditional behavior cloning, avoiding the need for value estimations. Typically, RvS \cite{emmons2021rvs} has investigated different conditional variables for MLP policy and establishes strong baseline for conditional behavior cloning. Many prior works are constructed in different styles of RvS, where the conditional variables can be the reward/return \cite{kumar2019reward,srivastava2019training,chen2021decision}, the target goal \cite{nair2018visual,emmons2021rvs,ghosh2021learning}, other task information \cite{codevilla2018end,hawke2020urban}, or trajectory-level aggregates \cite{furuta2021generalized,xie2023future}. 

Recent works \cite{chen2021decision,janner2021sequence} have explored another approach to perform conditional behavior cloning via sequence modeling by leveraging the Transformer architecture \cite{vaswani2017attention}, which has achieved widespread success in various machine learning fields, including natural language processing (NLP) \cite{radford2019language,devlin2018bert}, speech \cite{subakan2021attention,li2019neural}, and computer vision (CV) \cite{dosovitskiy2020image,carion2020end}. Specifically, Decision Transformer (DT) \cite{chen2021decision} utilizes a causal transformer architecture for the policy network to perform return-conditioned behavioral cloning, where instead of taking a single state as input, it takes a sequence of states, actions, and returns. Similarly, a model-based offline RL method is introduced in \cite{janner2021sequence}, namely Trajectory Transformer (TT), which also leverages the Transformer architecture to formulate the forward dynamics model and uses beam search for planning. Trajectory Autoencoding Planner \cite{jiang2023efficient} further enhances TT to achieve greater efficiency and scalability in high-dimensional action spaces. 

Building upon the achievements of DT, numerous studies have expanded its application to various contexts, such as online DT for online RL \cite{zheng2022online}, Q-learning DT for offline RL \cite{yamagata2023q}, or Prompt DT for few-shot learning \cite{xu2022prompting}. In the view of RvS, these works adhere to the reward conditioning style. However, recent works \cite{badrinath2024waypoint, brandfonbrener2022does, correia2023hierarchical} have explored the limitations of using reward as a conditional variable. Notably, DT has been identified as lacking in stitching ability \cite{badrinath2024waypoint}, a crucial property for offline RL algorithms to succeed on unstructured and suboptimal datasets, which often involve combining sub-trajectories from different demonstrations. Waypoint Transformer (WT) \cite{badrinath2024waypoint} addressed this issue by conditioning the DT on predicted $K$-step ahead goals instead of return-to-go, with these intermediate goals serving as future guidance for policy in goal-conditioned RL. In contrast to WT, we propose conditioning on predictive coding, which offers a more generalized and effective encoding of future information for guiding policy learning.

Masked autoencoding (MAE) has been demonstrated as an effective approach for acquiring representations in both NLP \cite{devlin2018bert, brown2020language} and CV \cite{he2022masked, bao2021beit}. In MAE, the input sequence undergoes random masking before being processed by a transformer encoder, resulting in a compressed representation that is subsequently fed into the transformer decoder to reconstruct the input. This straightforward yet effective approach has sparked recent advancements in RL \cite{carroll2022uni,wu2023masked,liu2022masked}. By leveraging masked autoencoder trained on randomly masked sequences to reconstruct the original data, these works have developed versatile models capable of performing a range of decision-making tasks, such as forward dynamics, inverse dynamics, behavior cloning, and state representations, by simply changing the masking pattern at inference time. In contrast to these approaches, we leverage the masked autoencoder to compress both the state-only trajectory and the desired goal, using the obtained compressed representation to guide policy learning. While our pretraining scheme draws inspiration from MaskDP \cite{liu2022masked} and MTM \cite{wu2023masked}, our focus is on addressing the challenges of DT in the goal-conditioned RL setting, particularly its stitching ability, rather than solely concentrating on representation learning.

\section{PRELIMINARIES}
\subsection{Goal-conditioned Reinforcment Learning}

A reinforcement learning (RL) environment is typically modeled by a Markov decision process (MDP), which can be described as a tuple of $\mathcal{M} = (\mathcal{S}, \mathcal{A}, R, P, \gamma)$. The tuple consists of states $s\in \mathcal{S}$, actions $a \in \mathcal{A}$, a reward function $r=R(s, a)$, transition dynamics distribution $P(s^{\prime}|s, a)$, and a discount factor $\gamma \in [0, 1)$. A trajectory is composed of a sequence of state-action pairs: $\tau=\{(s_t, a_t)\}_{t=1}^H$, where $H$ is the horizon, and $s_t, a_t$ refer to the state and action at timestep $t$, respectively. Additionally, we define $\tau_{i:j}=\{(s_t, a_t)\}_{t=i}^j$ as a sub-trajectory of $\tau$. An RL agent takes actions based on a policy $\pi(a|s)$. The objective is to maximize the expected return over trajectories $\mathbb{E}_{\pi}\left[\sum_{t=1}^{H}\gamma^{t - 1} R(s_t, a_t)\right]$. 

To extend RL to multiple tasks, a goal-conditioned formulation can be used \cite{kaelbling1993learning}, where the original MDP $\mathcal{M}$ is augmented with the goal space $\mathcal{G}$. We assume that the goal space $\mathcal{G}$ is the same as the state space (\ie, $\mathcal{G} \equiv \mathcal{S}$). Our goal is to learn an optimal goal-conditioned policy $\pi(a|s, g)$ that maximizes $\mathbb{E}_{g\sim p(g), \pi}\left[\sum_{t=1}^H\gamma^{t-1}R(s_t, a_t, g)\right]$, where the goal $g$ is sampled from a goal distribution $p(g)$. In this setting, a sparse goal-conditioned reward function is often used, where $R(s, a, g) = 1$ when $s = g$, and $0$ otherwise. 

\subsection{Offline RL via Supervised Sequence Modeling}
In offline RL, instead of obtaining data by interacting with the environment, we only have access to a fixed, limited dataset collected by unknown policies, such as human experts or baseline hand-engineered policies. This dataset contains a set of $N$ trajectories, $\mathcal{D} = \{\tau^i\}_{i=1}^N$, associated with rewards. We follow prior work that leverages sequence modeling for offline reinforcement learning \cite{chen2021decision}. Specifically, Decision Transformer (DT) \cite{chen2021decision} considers the following trajectory representation as input:
\begin{equation*}
    \small
    \hat{\tau} = (\hat{R}_1, s_1, a_1, \dots, \hat{R}_H,s_H, a_H),
\end{equation*}
where, $\hat{R}_t = \sum_{t'=t}^H r_{t'}$ represents the return-to-go, defined as the sum of future rewards from timestep $t$. Modality-specific encoders are employed to transform data (\ie, state, action, and return-to-go) from the raw modality space to a common representation space, to which timestep embeddings are additionally added. After tokenization, $\hat{\tau}$ is input into a GPT-based transformer \cite{radford2018improving}, serving as an expressive policy function approximator $\pi_{\theta}$ to predict the next actions. The policy is trained to maximize the likelihood of actions within the dataset:
\begin{equation}
    \small
    \mathcal{L}_{\text{DT}} = \mathbb{E}_{\hat{\tau} \sim \mathcal{D}}\left[-\sum_t\log \pi_{\theta}(a_t|\hat{\tau}_{t-k + 1:t-1}, s_t, \hat{R}_t)\right]
\end{equation}
At training time, return-to-go sequences are computed from offline trajectories. During inference, this quantity guides the policy in generating actions to achieve the desired outcome. However, using return-to-go as a conditional variable presents certain disadvantages. First, at test time, the full trajectory is unknown, and its length may vary, making it difficult to calculate the sum. Consequently, users must specify the target return heuristically. Moreover, the performance of DT is shown to be sensitive to this value \cite{correia2023hierarchical,yang2023dichotomy}. Second, in sparse-reward tasks where rewards are almost constant (\eg, $r_t = 0$) until reaching the final target, return-to-go values remain static. This challenge impedes DT's ability to learn tasks effectively, especially in long-horizon tasks \cite{correia2023hierarchical}. Third, using a single scalar value as input may fail to capture sufficient future information. These limitations motivate the design of new, effective transformer-based algorithms.

\begin{figure}[t]
    \centering
    \includegraphics[width=0.49\textwidth]{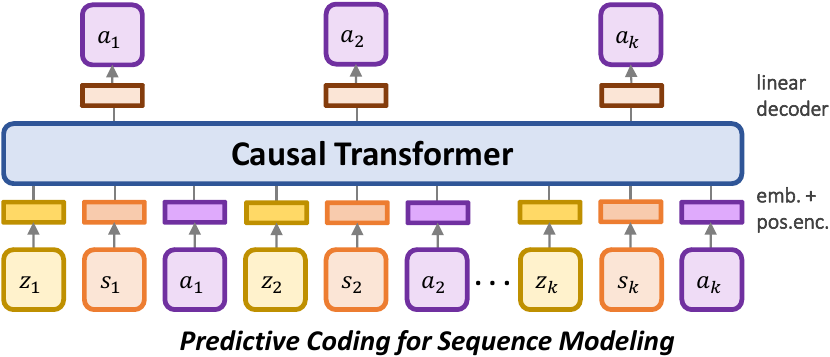}
    \vskip -0.05in
    \caption{The overview of the \ours \ model. \ours \ takes as input a length-$k$ sub-trajectory associated with predictive latent codings to make decisions. During training, predictive codings are extracted from states within the same sub-trajectory. During inference, \ours \ generates the next action in an autoregressive manner, akin to DT  \cite{chen2021decision}.}
    \label{fig:pedt}
    \vskip -0.2in
\end{figure}

\section{PROPOSED APPROACH}

To eliminate the reliance on reward information, inspired by future-conditioned RL \cite{furuta2021generalized,emmons2021rvs}, we introduce Predictive Coding for Decision Transformer (\ours), a framework that enables policy learning guided by future information. 

\subsection{Policy Learning based on Predictive Coding} \label{sec:pcdt}

The overview of \ours \ is presented in Fig. \ref{fig:pedt}. Let $f_{\phi}^E(\cdot)$ be a trajectory encoder, and $\pi_{\theta}(\cdot)$ be a policy network. In this framework, our objective is to condition the policy on predictive coding $z$ instead of the return, resulting in an alternative representation of the trajectory input:
\begin{equation*}
    \small
    \bar{\tau} = (z_1, s_1, a_1, \dots, z_H, s_H, a_H),
\end{equation*}
where, the latent variables are obtained by $z_{t-k+1:t} = f_{\phi}^E(s_{t-k+1:t}, g)$, as depicted in Fig. \ref{fig:trajectory_enc}. These latent variables $z_i$ are expected to capture not only the historical context and task information (\ie, the desired goal) but also to be predictive, enabling reasoning over future states toward the desired goal. During inference, given past states and actions, \ours \ utilizes the learned policy $\pi_{\theta}$ along with the learned trajectory encoder $f_{\phi}^E$ to autoregressively produce the action:
\begin{equation*}
    \small
    a_t \leftarrow \pi_{\theta}(\cdot |\bar{\tau}_{t-k+1:t-1}, s_t, z_t). 
\end{equation*}
Note that, both the policy and trajectory encoder take the same view of the $k$-length input sequence, \ie, from timestep $t-k+1$ to $t$.

Similar to DT, we leverage the GPT-based Transformer \cite{radford2018improving} to parameterize the policy network $\pi_{\theta}$. However, we opt for sinusoidal positional encoding over timestep encoding to mitigate overfitting \cite{carroll2022uni}. Given sub-trajectories sampled from the dataset $\mathcal{D}$, we first compute the corresponding latent variables $z_i$, then the policy network is trained by minimizing the behavioral cloning objective as follows:
\begin{equation}\label{eq:pcdt_loss}
    \small
    \mathcal{L}(\theta) = \mathbb{E}_{\bar{\tau} \sim \mathcal{D}}\left[-\sum_{t}\log\pi_{\theta}(a_t|\bar{\tau}_{t-k+1:t-1}, s_t, \lfloor z_t \rfloor)\right]
\end{equation}
where, $\lfloor \cdot \rfloor$ denotes the stop-gradient operator. Note that during policy training, the gradient only backpropagates through the policy network. The trajectory encoder is learned by a separate learning objective, which we describe in the next section. Utilizing predictive codings as conditioning offers several advantages. Firstly, it avoids the need for reward signals during learning, thereby enabling expansive, large-scale training opportunities. Secondly, it mitigates the issue of inconsistent behaviors stemming from return-conditioned behavioral cloning during testing, where behaviors often deviate significantly from the desired target \cite{correia2023hierarchical,yang2023dichotomy}.

\subsection{Predictive Coding Learning} \label{sec:predictive_coding}
\begin{figure}[t]
    \centering
    \includegraphics[width=0.49\textwidth]{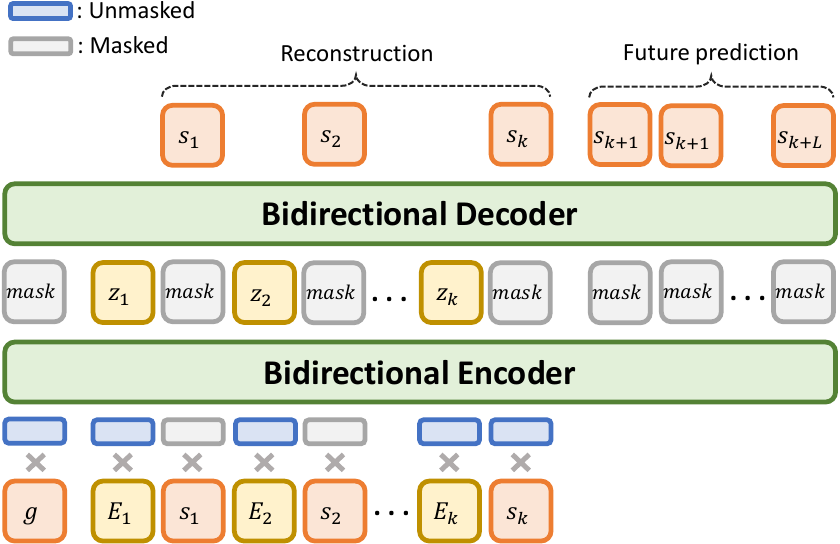}
    \caption{We input the sequence of state-dummy token pairs concatenated with the target goal into the trajectory encoder. Within this sequence, states are randomly masked. The trajectory decoder receives stacked inputs of latent codes and masked tokens to reconstruct the input states. Additionally, in addition to reconstruction, we leverage the same latent codes for predicting future states, aiming to enhance predictiveness.}
    \label{fig:trajectory_enc}
    \vskip -0.2in
\end{figure}

In order to condense the trajectory into a compact latent space, we utilize masked autoencoding with both the encoder and decoder parameterized by a series of bidirectional transformers and trained via masked prediction, similar to previous works \cite{carroll2022uni, liu2022masked, wu2023masked}. An overview of the trajectory autoencoder is shown in Fig. \ref{fig:trajectory_enc}. We use $f_{\phi}^E$ and $f_{\phi}^D$ to represent the trajectory encoder and decoder, respectively.

Specifically, the trajectory encoder takes as input the concatenation of \textit{the target goal} $g$, the dummy tokens $E_{1:k}$, and the length-$k$ sequence of states $s_{t-k+1:t}$. The dummy tokens are shared across input sequences. During training, the states are randomly masked to improve generalization, while the goal and dummy tokens remain unmasked, as illustrated in Fig. \ref{fig:trajectory_enc}. The goal $g$ is uniformly sampled from reachable states $s_{t:H}$ within the same trajectory. These input tokens are then encoded by modality-specific encoders, parameterized by a linear layer, with sinusoidal positional embeddings added. Subsequently, the embeddings are processed by the bidirectional transformer to yield latent codes $z$. It's worth noting that only unmasked tokens are provided to the transformer, mirroring the approach of MAE-based methods \cite{he2022masked, carroll2022uni, liu2022masked, wu2023masked}. To reconstruct the input states, the trajectory decoder takes as input the latent code tokens corresponding to the position of dummy tokens, the remaining tokens are substituted by masked tokens. To encourage predictive coding, in addition to reconstructing input states, future state prediction is concurrently performed from the latent codes. The encoder and decoder are jointly trained by minimizing the following learning objective:
\begin{equation}\label{eq:trajectory_loss}
    \small
    \mathcal{L}(\phi) = \mathbb{E}_{s, g \sim \mathcal{D}}\left[\sum_{t}\left(f_{\phi}(s_{t-k+1:t}, g) - s_{t-k+1:t+L}\right)^2\right]
\end{equation}
where, $f_{\phi}$ represents both the encoder and decoder. Also, we consider dummy tokens $E_{1:k}$ as part of the encoder, thus omitting them from the input of $f_{\phi}$. The ground truth states include $k$ historical states and $L$ future states, \ie, from timestep $t-k+1$ to $t+L$. During policy learning and inference, only the trajectory encoder is used, applied to the unmasked observed states and the target goal to acquire predictive codings. With this design, we can learn the predictive codes to perform goal-conditioned future prediction. Consequently, these predictive codes serve as guidance toward the desired goal while being aware of future behaviors during policy learning. Besides, we only leverage states from the dataset for training the predictive coding, enabling us to utilize a potentially large amount of action-free data \cite{xu2022policy,ghosh2023reinforcement}. We leave this for future work. The procedure for training trajectory autoencoder and policy is summarized in Algorithm \ref{alg:pcdt_algorithm}.
\begin{algorithm}[t]
    \caption{\ours \ training algorithm.}
    \label{alg:pcdt_algorithm}
    \begin{algorithmic}[1]
        \STATE \textbf{Input}: Demonstration $\mathcal{D} = \{\tau_i\}_{i=1}^N$, length of historical states $k$, length of future states $L$, and learning rate $\alpha$.
        \STATE \textbf{Initialize}: trajectory autoencoder $f_{\phi}$ and policy $\pi_{\theta}$.
        \STATE // \texttt{Training the trajectory autoencoder}
        \WHILE{not converged}
            \STATE Randomly sample trajectories, $\tau \sim \mathcal{D}$
            
            \STATE Sample timestep for each trajectory, $t \sim [1, H]$, obtain $s_{t-k+1:t+L}$, and sample goal, $g \sim s_{t:H}$

            \STATE Calculate objective function $\mathcal{L}(\phi)$ as in Eq. (\ref{eq:trajectory_loss})
            
            \STATE Update autoencoder parameters: $\phi \leftarrow \phi + \alpha \nabla_{\phi}\mathcal{L}(\phi)$
        \ENDWHILE
    
        \STATE // \texttt{Training the policy}
        \WHILE {not converged}
            \STATE Randomly sample trajectories, $\tau \sim \mathcal{D}$
            
            \STATE Sample timestep for each trajectory, $t \sim [1, H]$, obtain $s_{t-k+1:t}$ and $a_{t-k+1:t}$, and sample goal, $g \sim s_{t:H}$
            
            \STATE Compute predictive codings:\\
            \begin{center}
                $z_{t-k+1:t} = f_{\phi}^E(s_{t-k+1:t}, g)$    
            \end{center}

            \STATE Calculate objective function $\mathcal{L}(\theta)$ as in Eq. (\ref{eq:pcdt_loss})
            
            \STATE Update policy parameters: $\theta \leftarrow \theta + \alpha \nabla_{\theta}\mathcal{L}(\theta)$
            
        \ENDWHILE
    \end{algorithmic}
\end{algorithm}
\section{EXPERIMENTS}

\subsection{Experimental Setup}

\begin{table*}[t]
\renewcommand{\arraystretch}{1.2}
\centering
\caption{Average normalized scores of \ours \ against other baselines on AntMaze and FrankaKitchen from D4RL benchmark \cite{fu2020d4rl}. Following \cite{kostrikov2021offline}, we bold all scores within 5 percent of the maximum per dataset ($\geq$ 0.95 $\cdot$ max).}
\label{tab:comparison}
\begin{tabular}{l|c@{\hspace{5pt}}cc|c@{\hspace{5pt}}ccccc}

\noalign{\hrule height 0.8pt}
\textbf{Dataset}  
& \textbf{CQL}  & \textbf{IQL}      & \textbf{GC-IQL} 
& \textbf{BC}   & \textbf{GCBC}   & \textbf{DT}$_{\textbf{R}}$  & \textbf{DT}$_{\textbf{G}}$  & \textbf{WT}   &  \textbf{PCDT} (Ours)  \\
\noalign{\hrule height 0.8pt}
antmaze-umaze   
& $ 74.0 $  & $ \bf 87.5 \pm 2.6 $  & -
& $ 54.6 $  & $ 65.6 \pm 9.9 $   & $ 53.6 \pm 7.3 $  & $ 61.2 \pm 5.8 $  &  $ 64.9 \pm 6.1 $  &  $ 70.8\pm 3.9$  \\
antmaze-umaze-diverse   
& $ \bf 84.0 $  & $ 62.2\pm 13.8$   & -
& $ 45.6 $  & $ 60.9\pm 11.2$   & $ 42.2 \pm 5.4$   & $ 66.0 \pm 5.3$   & $ 71.5 \pm 7.6 $   &  $ 71.5 \pm 3.0$  \\
antmaze-medium-play   
& $61.2$    & $ 71.2\pm 7.3$    & $ 70.9 \pm 11.2 $
&  $ 0.0 $  & $ \bf 71.9\pm 16.2$   & $ 0.0 \pm 0.0 $   & $ 56.0 \pm 9.4$ & $ 62.8 \pm 5.8 $  &  $ \bf 75.2\pm 5.7$  \\
antmaze-medium-diverse   
& $53.7$    & $ 70.0\pm 10.9$   & $ 63.5 \pm 14.6 $     
&  $ 0.0 $  & $ 67.3\pm 10.1$   & $ 0.0 \pm 0.0 $   & $ 59.5  \pm 7.6$ & $ 66.7 \pm 3.9 $  &  $ \bf 83.2\pm 4.8$  \\
antmaze-large-play   
& $15.8$    & $ 39.6\pm 5.8$    & $ 56.5 \pm 14.4 $
&  $ 0.0 $  & $ 23.1\pm 15.6$   & $ 0.0 \pm 0.0 $   & $ 22.0  \pm 7.5 $ & $ \bf 72.5 \pm 2.8 $  &  $ \bf 74.8 \pm 6.2$  \\
antmaze-large-diverse   
& $14.9$    & $ 47.5\pm 9.5$    & $ 50.7 \pm 18.8 $
&  $ 0.0 $  & $ 20.2\pm 9.1$   & $ 0.0 \pm 0.0 $   &  $ 21.0 \pm 6.6 $ & $ \bf 72.0 \pm 3.4 $  &  $ \bf 73.6\pm 5.7$  \\
\noalign{\hrule height 0.8pt}
antmaze average
& $50.6$    & $ 63.0$           & $ 60.4 $
&  $ 16.7 $ & $ 51.5 $   & $ 16.0 $  &  $ 47.6 $  &  $ 68.4$ &  $ \bf 74.9 $ \\
\noalign{\hrule height 0.8pt}
kitchen-partial
& $49.8$    & $ 46.3$           & $ 39.2 \pm 13.5 $
&  $ 38.0 $ & $ 38.5\pm 11.8$   & $ 31.4 \pm 19.5 $ & $ 65.2 \pm 1.9 $  &  $ 63.8\pm 3.5$ &  $ \bf 74.5 \pm 5.0 $ \\
kitchen-mixed
& $51.0$    & $ 51.0$           & $ 51.3 \pm 12.8 $
&  $ 51.5 $ & $ 46.7\pm 20.1$   & $ 25.8 \pm 5.0 $  &  $ 55.4 \pm 3.8$  &  $ 70.9\pm 2.1$ &  $ \bf 75.6 \pm 4.7 $ \\
\noalign{\hrule height 0.8pt}
kitchen average
& $50.4$    & $ 48.7$           & $ 45.3 $ 
& $ 44.8 $  & $ 42.6$   & $ 28.6 $          &  $ 60.3 $  &  $ 67.4$ &  $ \bf 75.1 $ \\
\noalign{\hrule height 0.8pt}
average
& $50.6$    & $ 59.4$           & $ 54.6 $ 
& $ 23.7 $  & $ 49.3$   & $ 19.1 $          &  $ 50.8 $  &  $ 68.1$ &  $ \bf 74.9 $ \\
\noalign{\hrule height 0.8pt}
\end{tabular}
\vskip -0.15in
\end{table*}

\begin{figure}[t]
    \centering
    \includegraphics[width=2.75cm, height=2.75cm]{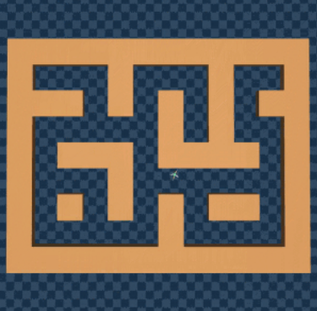}
    \includegraphics[width=2.75cm, height=2.75cm]{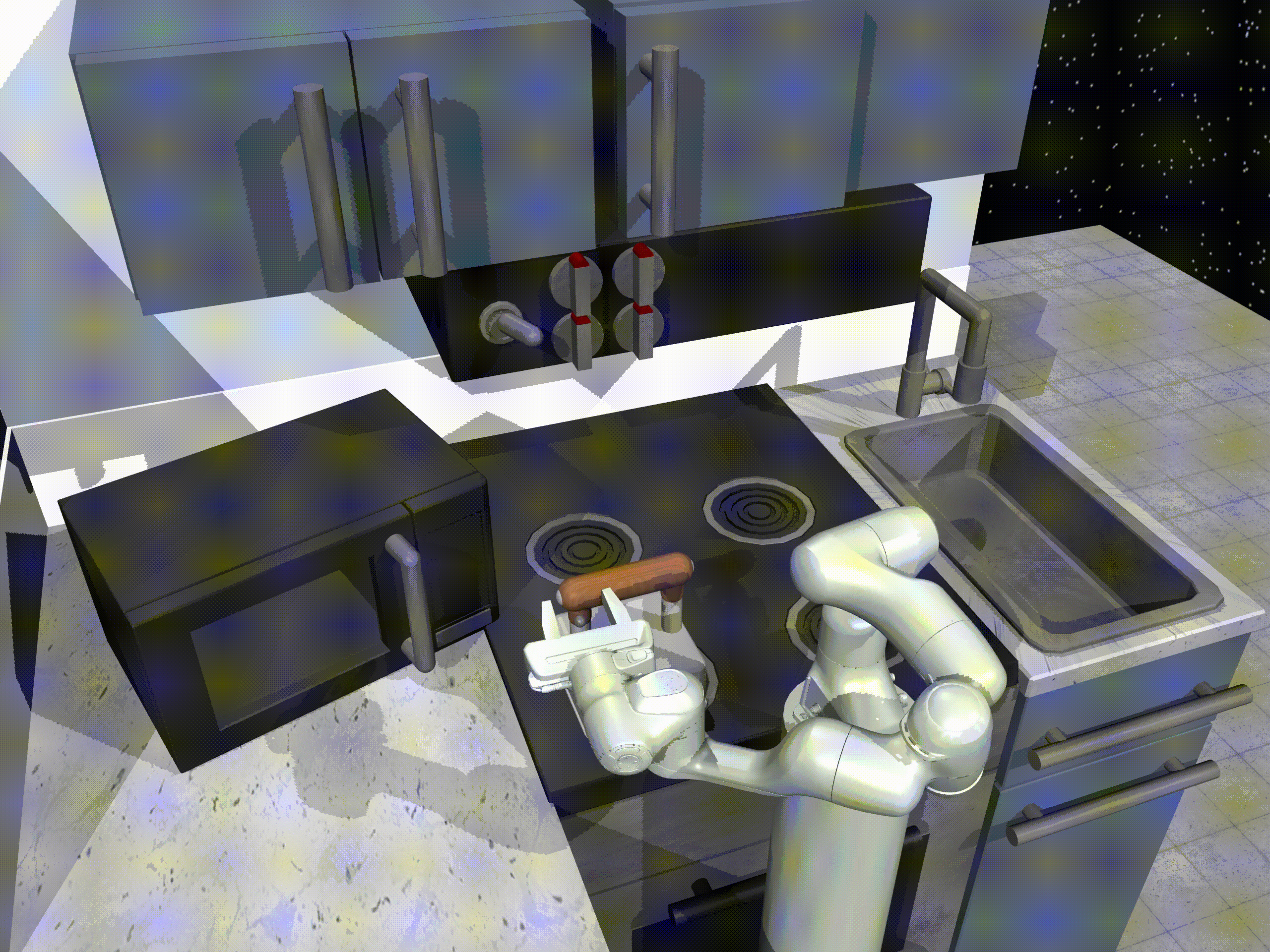}
    \includegraphics[width=2.75cm, height=2.75cm]{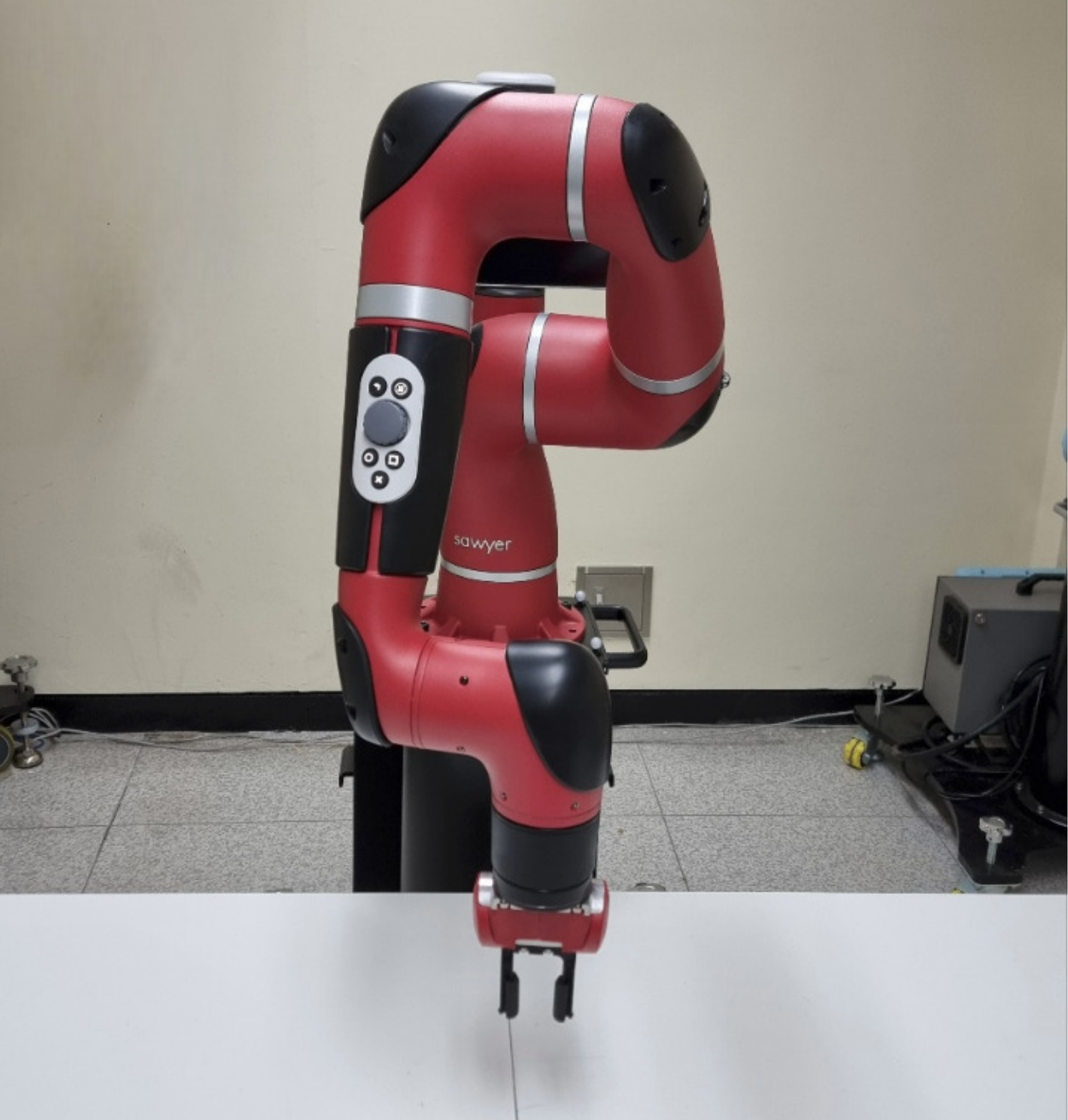}
    \caption{We evaluate the performance of \ours \ on three sparse-reward goal-conditioned tasks. From left to right: AntMaze and FrankaKitchen, taken from the D4RL benchmark \cite{fu2020d4rl}, and a physical Rethink Sawyer robot performing goal-reaching task.}
    \label{fig:environment}
    \vskip -0.1in
\end{figure}

In this section, we evaluate \ours \ on goal-conditioning benchmarks, specifically AntMaze, FrankaKitchen from D4RL \cite{fu2020d4rl} and a Rethinking Sawyer robot performing a goal-reaching task, as illustrated in Fig. \ref{fig:environment}. These environments pose a challenge for offline RL methods due to the scarcity of optimal trajectories and serve as rigorous benchmarks for evaluating a model’s stitching ability \cite{fu2020d4rl}. AntMaze comprises long-horizon navigation tasks with sparse rewards, involving the control of an 8-DoF Ant robot to navigate from its initial position to a specified goal location. Meanwhile, FrankaKitchen presents long-horizon manipulation tasks, where the objective is to complete four subtasks (\eg, closing the microwave or sliding open the cabinet) using a 9-DoF Franka robot. We train our algorithms on diverse datasets provided by these environments. Each dataset comprises suboptimal and undirected data, where the desired targets in demonstrations are unrelated to the evaluation tasks. Additionally, we also validate the effectiveness of our method on the real-world environment by conducting goal-reaching tasks with a 7-DOF Sawyer robot arm. By this, we provide concrete evidence of its viability and robustness in navigating real environments.

For the trajectory autoencoder, we employ a two-layer bidirectional transformer for the encoder and a one-layer bidirectional transformer for the decoder, each with four self-attention heads. The length of future states ($L$) in training the trajectory autoencoder is set to 100 for AntMaze and 40 for FrankaKitchen. For \ours, we utilize a causal transformer with the number of layers searched within $\{3, 4, 5\}$, each with one self-attention head. All transformer use hidden size of 256. Both training phases use the Adam optimizer with a learning rate of $1e-4$, a batch size of 1024, and 80 epochs. The input sequence length is set to 10 for AntMaze and 5 for FrankaKitchen. Following \cite{kostrikov2021offline}, we average over 100 episodes for AntMaze and 50 episodes for FrankaKitchen for each evaluation run. Reported scores include the mean and standard deviation over five seeds for each experiment. 

\subsection{Comparison with Previous Approaches}

\begin{figure}[t]
    \centering
    \includegraphics[width=0.9\columnwidth]{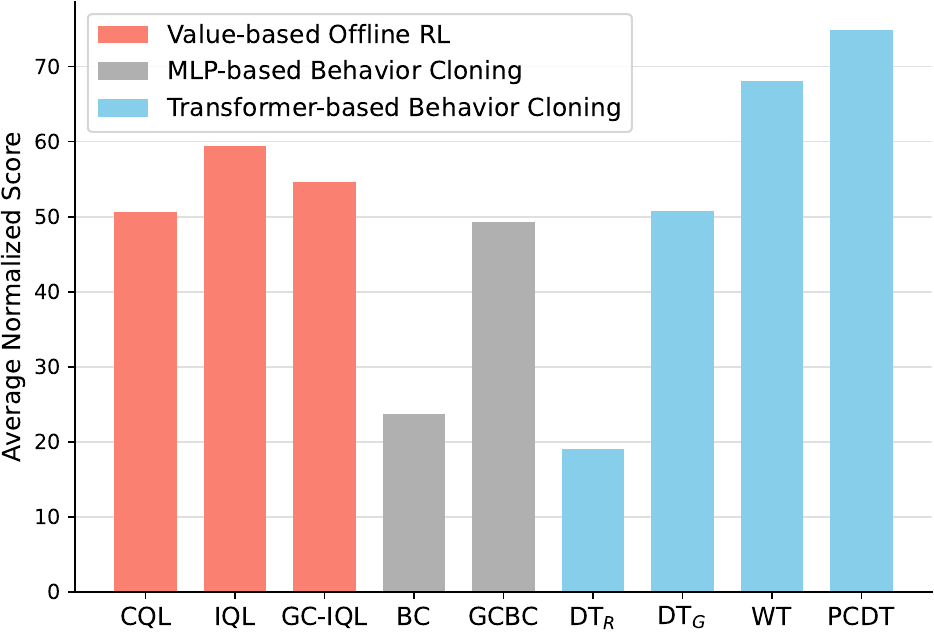}
    \caption{A plot illustrating the average performance of each algorithm listed in Table \ref{tab:comparison}. The bars are color-coded based on a simplified algorithm categorization.}
    \label{fig:comparison_category}
    \vskip -0.2in
\end{figure}

We compare the performance of \ours \ with both value-based offline RL and behavior cloning methods. Among the value-based techniques, we include CQL \cite{kumar2020conservative}, IQL \cite{kostrikov2021offline}, and its goal-conditioned variant, GC-IQL \cite{park2024hiql}. For the MLP-based behavior cloning approach, we evaluate BC and its goal-conditioned counterpart, GCBC \cite{ding2019goal,emmons2021rvs}, both of which utilize MLPs to parameterize the policy network. For transformer-based policies, we consider DT$_R$ \cite{chen2021decision}, which conditions on return-to-go, DT$_G$, which conditions on the target goal, and WT \cite{badrinath2024waypoint}, similar to our approach but conditioned on predicted $K$-step ahead goals. For all methods except DT$_G$, we use reported results from \cite{badrinath2024waypoint} and \cite{park2024hiql}. For DT$_G$, we modify DT by replacing return-to-go with the target goal using the official implementation provided by \cite{chen2021decision}.

Table \ref{tab:comparison} and Fig. \ref{fig:comparison_category} present the results across eight offline datasets. When compared to value-based methods, \ours \ outperforms six out of eight datasets. Across all datasets, \ours \ achieves a score of 74.9, representing a substantial improvement over the top-performing value-based method, IQL (59.4), with a relative percentage increase of 26.1\%. Notably, in the most challenging datasets requiring stitching like AntMaze Large, our method exhibits remarkable enhancements, with relative improvements of 88.9\% (``play'') and 54.9\% (``diverse'') over IQL. Similarly, in the FrankaKitchen Mixed and Partial, we observe substantial enhancements of 60.9\% and 48.2\%, respectively. These results demonstrate the efficacy of our predictive codings in generalizing behaviors from suboptimal trajectories and effectively addressing the stitching problem encountered by DT. 

For the behavior cloning baselines, we find that our method performs on par with or better than all the previous methods. In particular, \ours \ exhibits significant improvements over DT$_R$ by a factor of 3.x in average performance. This notable improvement underscores the efficacy of leveraging predictive codings, which offer a stronger signal for policy learning compared to the return sequence. Moreover, by naively conditioning on target goals, DT$_G$ demonstrates some learning capability across AntMaze datasets, which again emphasizes the limitations of using rewards as conditional variables in sparse-reward goal-conditioned RL. Despite its efforts, the performance of DT$_G$ remains considerably distant from \ours, with ours showcasing a substantial 47.4\% improvement in average performance over DT$_G$. In comparison with the prior state-of-the-art, our method improves over WT by 9.5\% and 11.4\% on AntMaze and FrankaKitchen, respectively. These results underscore the ability of predictive codings to encode more comprehensive future information compared to intermediate future goals, thereby more effectively guiding the policy toward the desired target.

\begin{figure}[t]
    \centering
    \includegraphics[width=0.49\columnwidth]{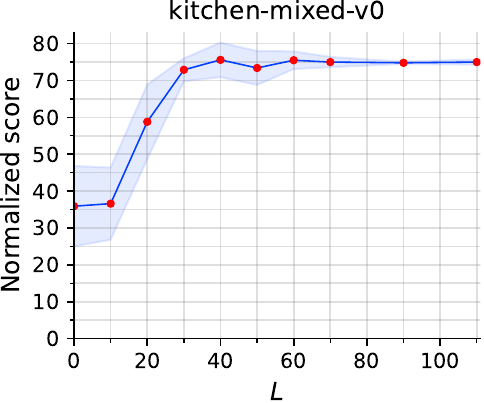}
    \includegraphics[width=0.49\columnwidth]{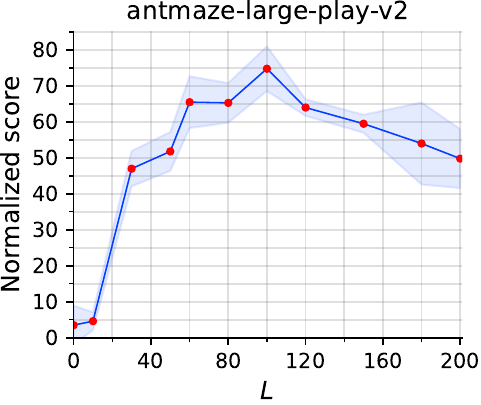}
    \vskip -0.05in
    \caption{The effect of different lengths of future states ($L$) during the learning of predictive coding on agent performance.}
    \label{fig:ablate_future_horizon}
    \vskip -0.2in
\end{figure}

\subsection{Ablation Studies} 

\textbf{Effectiveness of Future State Prediction.} We first investigate how future prediction impacts the quality of learned predictive coding. For this analysis, we consider \textit{antmaze-large-play-v2} and \textit{kitchen-mixed-v0}, two tasks crucial for assessing the stitching ability of offline RL algorithms. To elucidate the influence of predictiveness on policy learning, we vary the $L$-length future states when training trajectory encoders (see Section \ref{sec:predictive_coding}) and evaluate the agent's performance based on the learned predictive codings. Here, $L = 0$ indicates that the trajectory autoencoder solely reconstructs the input sequence without considering future prediction. The agent's normalized score is plotted against $L$ in Fig. \ref{fig:ablate_future_horizon}. In AntMaze Large, an optimal $L$ is around 100, while in FrankaKitchen, $L\geq 30$ yields good performance. Importantly, when $L$ is too small, \eg, $\leq 10$ in both cases, the agent's performance drastically drops compared to other $L$ values; the score decreases by a factor of 2.1$\times$ and 16.3$\times$ in FrankaKitchen and AntMaze Large, respectively, compared to the optimal $L$. This analysis reveals that incorporating future information through predictive coding can markedly enhance task-solving capabilities, empowering the agent to anticipate future behaviors.

\begin{table}[t]
\centering
\caption{We ablate transformer configurations during policy learning, including dropout probability ($p_{\text{drop}}$) and number of transformer layers ($N$). The highest score is highlighted in bold and used in the main results.}
\label{tab:ablation_param}
\begin{tabular}{l@{\hspace{6pt}}ccc}
\toprule
$p_{\text{drop}}$ & antmaze-medium-diverse & antmaze-large-play & kitchen-mixed \\ 
\midrule
0.00 & $ 74.0\pm8.8 $ & $ 68.0 \pm 10 $ & $ 72.0 \pm 2.3 $   \\
0.05 & $ 73.2\pm5.2 $ & $ 70.7 \pm 6.4 $ & $ \bf 75.6 \pm 4.7 $  \\
0.10 & $ \bf 83.2\pm4.8 $ & $ \bf 74.8 \pm 6.2 $ & $ 70.0 \pm 5.2 $  \\
0.15 & $ 80.4\pm7.1 $ & $ 62.5 \pm 3.4 $ & $ 71.0 \pm 2.9 $  \\
0.20 & $ 76.0\pm5.2 $ & $ 59.8 \pm 4.9 $ & $ 69.8 \pm 2.2 $  \\
0.25 & $ 70.0\pm5.2 $ & $ 48.9 \pm 8.4 $ & $ 66.0 \pm 7.1 $  \\
\midrule
\midrule
$N$ & antmaze-medium-diverse & antmaze-large-play & kitchen-mixed \\ 
\midrule
1 & $ 70.8\pm3.9 $ & $ 24.0 \pm 1.8 $ & $ 52.8 \pm 7.9 $  \\
2 & $ 74.0\pm5.8 $ & $ 49.5 \pm 5.7 $ & $ 62.3 \pm 3.9 $  \\
3 & $ \bf 83.2\pm4.8 $ & $ 61.5 \pm 6.3 $ & $ 52.3 \pm 6.0 $  \\
4 & $ 72.4\pm6.5 $ & $ \bf 74.8 \pm 6.2 $ & $ 70.1 \pm 5.1 $  \\
5 & $ 72.0\pm3.1 $ & $ 63.3 \pm 8.1 $ & $ \bf 75.6 \pm 4.7 $  \\
6 & $ 75.2\pm4.3 $ & $ 58.5 \pm 8.4 $ & $ 70.3 \pm 4.9 $  \\
7 & $ 64.0\pm5.8 $ & $ 58.0 \pm 3.9 $ & $ 73.8 \pm 1.4 $  \\
\bottomrule
\end{tabular}
\vskip -0.25in
\end{table}

\textbf{Capacity and Regularization.} Following the work in \cite{emmons2021rvs}, we balance between capacity and regularization to maximize policy performance. We explore transformer configurations during policy learning, specifically, ablation of the dropout probability $p_{\text{drop}}$ and the number of transformer layers $N$. For this investigation, we consider the \textit{antmaze-medium-diverse-v2}, \textit{antmaze-large-play-v2}, and \textit{kitchen-mixed-v0} datasets. Based on the results in Table \ref{tab:ablation_param}, we observe that sensitivity to the various ablated hyperparameters is relatively low in terms of performance on \textit{antmaze-medium-diverse-v2} and \textit{kitchen-mixed-v0}, with a decrease by a factor of 1.1-1.4 compared to the best hyperparameter. However, in \textit{antmaze-large-play-v2}, we observe that choosing the appropriate number of layers matters for the agent's performance, where $N=1$ decreases performance by a factor of 3.x compared to $N=4$.

\begin{figure}[t]
    \centering
    \begin{overpic}[width=0.48\columnwidth]{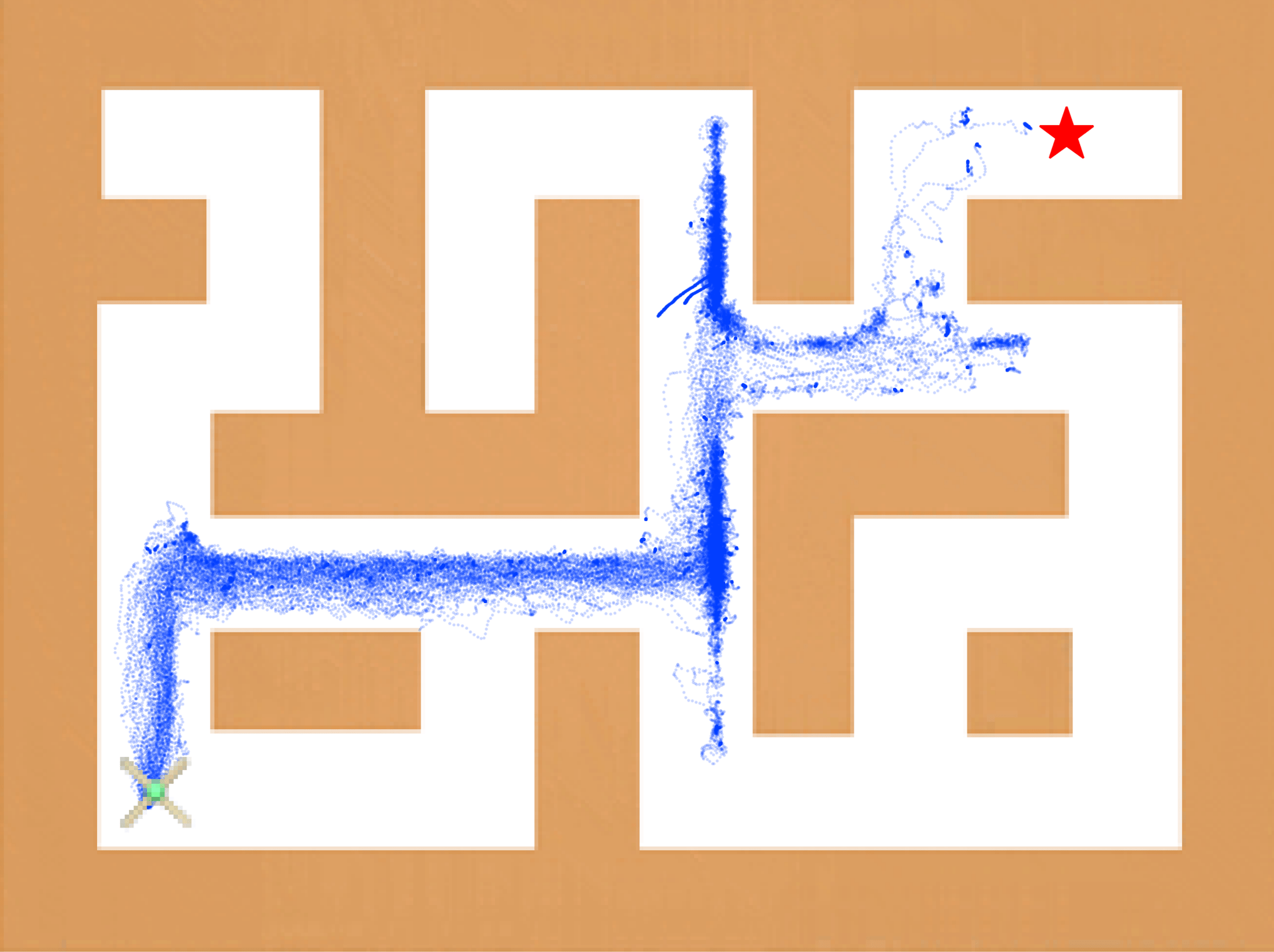}
        \put(0,69.3){\footnotesize\bfseries(a)}
    \end{overpic}
    \hspace{0.01\columnwidth} 
    \begin{overpic}[width=0.48\columnwidth]{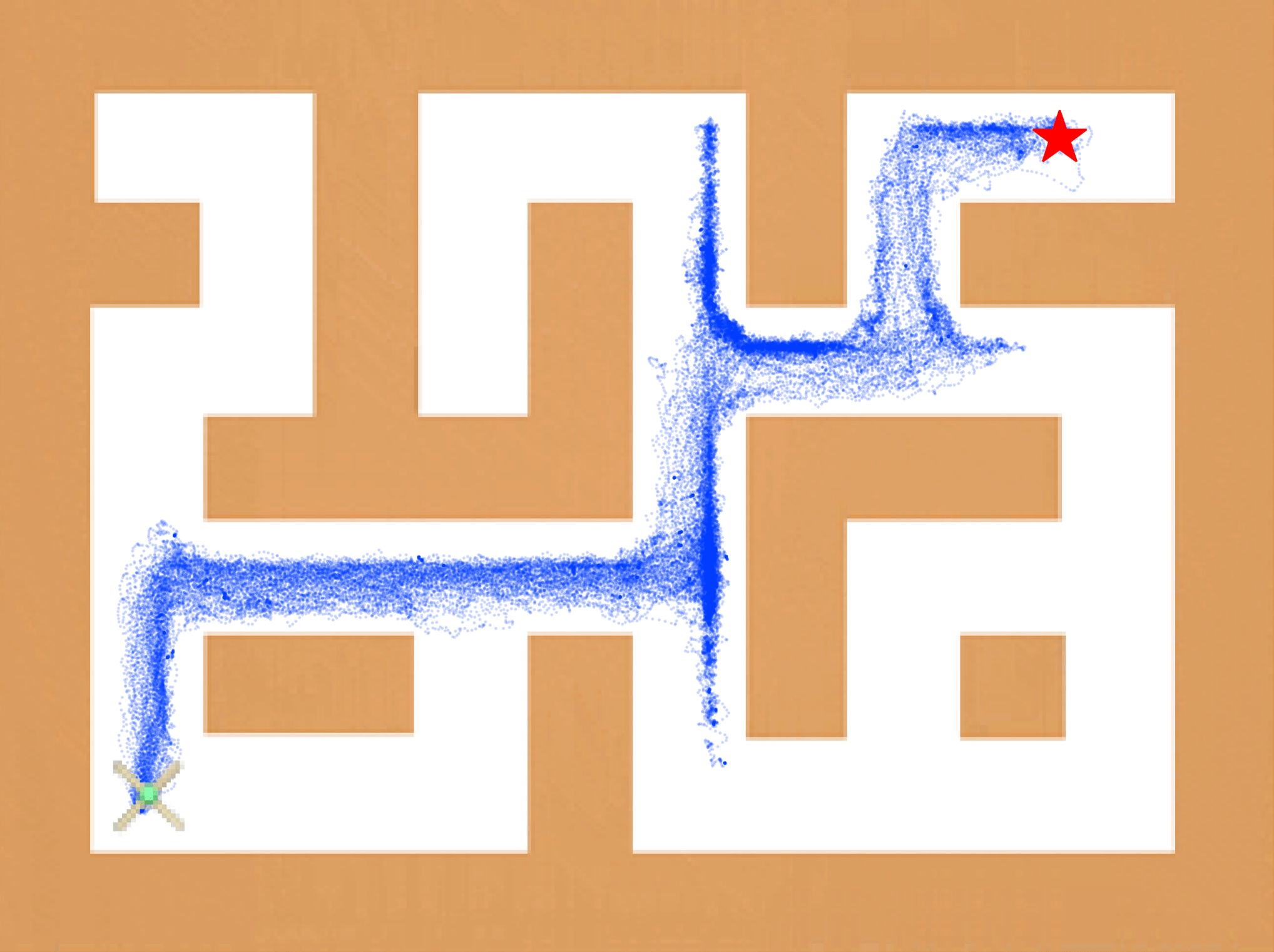}
        \put(0,69.3){\footnotesize\bfseries(b)}
    \end{overpic}\\
    \vspace{0.1cm} 
    \begin{overpic}[width=0.48\columnwidth]{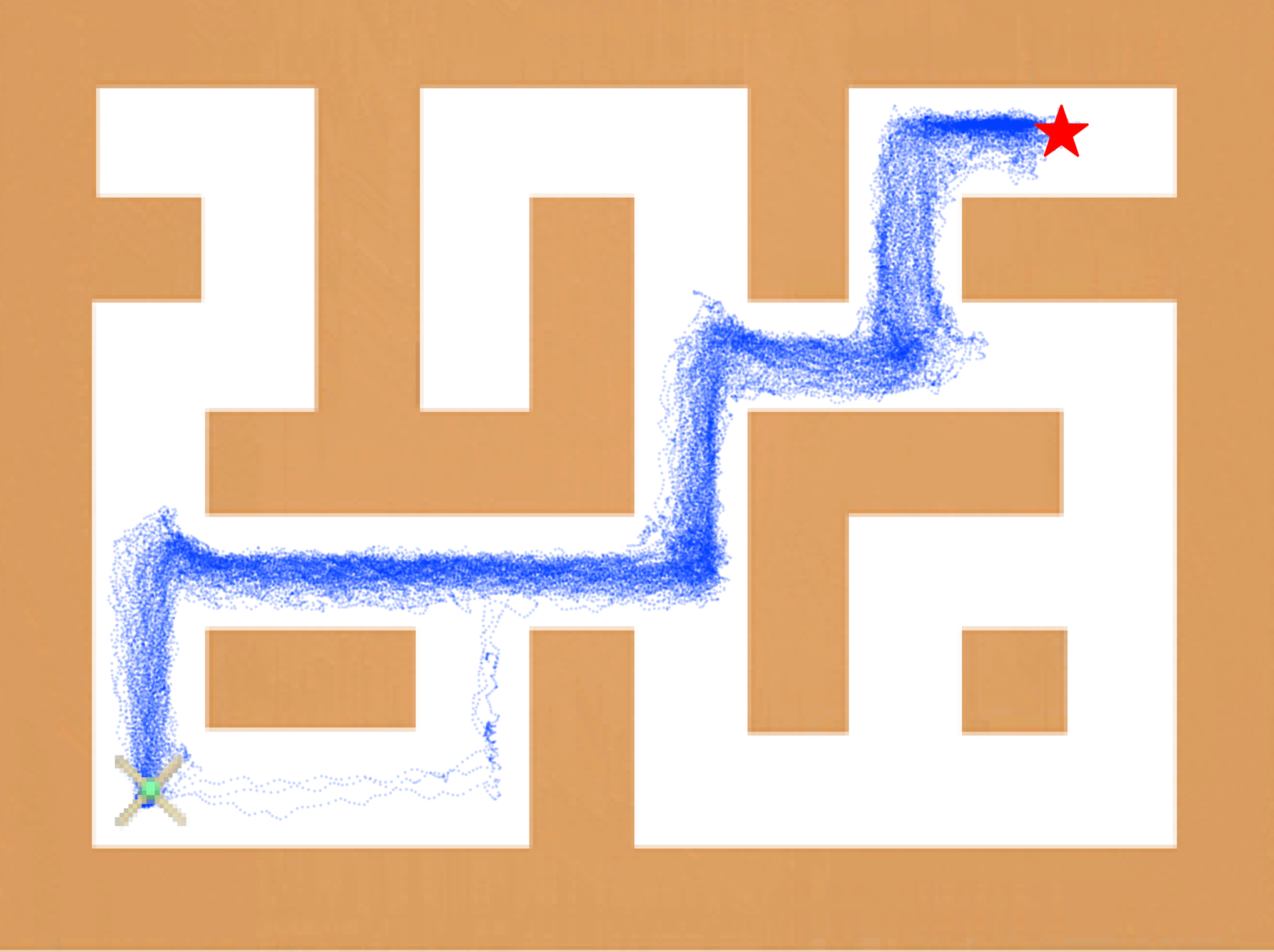}
        \put(0,69.3){\footnotesize\bfseries(c)}
    \end{overpic}
    \hspace{0.01\columnwidth} 
    \raisebox{-3.5mm}{\begin{overpic}[width=0.48\columnwidth]{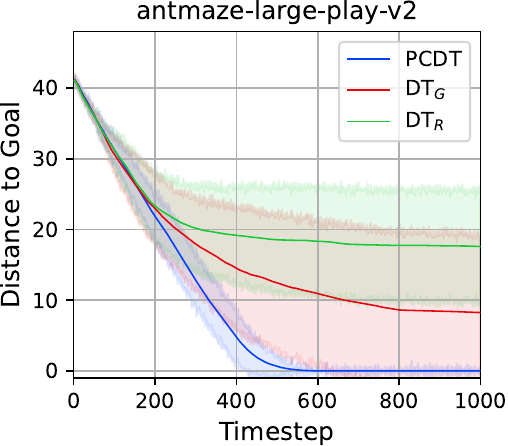}
        \put(15,76){\footnotesize\bfseries(d)}
    \end{overpic}}
    \caption{Visualization of 100 trajectories from the policies trained on \textit{antmaze-large-play-v2}: (a) DT$_R$ policy, (b) DT$_G$ policy, and (c) \ours \ policy; (d) the distance from the ant's location to the target goal position at each timestep.}
    \label{fig:antmaze_visualization}
    \vskip -0.15in
\end{figure}

\textbf{Qualitative Result.} To gain insights into the agent's behavior under the guidance of predictive codings, we qualitatively evaluate the performance across rollouts of trained policies in the \textit{antmaze-large-play-v2} dataset. Specifically, we examined the performance of three transformer-based agents: DT$_R$, DT$_G$, and \ours. The visualization of the ant's location across 100 trajectories for each agent is presented in Fig. \ref{fig:antmaze_visualization}. Our observations indicate that DT$_R$ (Fig. \ref{fig:antmaze_visualization}a), guided solely by return-to-go, struggles to consistently reach the desired target. While DT$_G$ (Fig. \ref{fig:antmaze_visualization}b), guided by target goals, occasionally reaches the target, the ant's behaviors exhibit inconsistencies, often veering off course in the middle of the map. In contrast, \ours \ (Fig. \ref{fig:antmaze_visualization}c) demonstrates a higher level of ability and consistency in reaching the goal location. Additionally, it sometimes reaches the target via alternative routes (\eg, bottom left of Fig. \ref{fig:antmaze_visualization}c), indicating that our method does not solely overfit to the dataset but also has the capability of generalization. As a result, \ours \ achieves a significantly higher score than DT$_G$ by about 3 times and takes fewer steps to complete the task (Fig. \ref{fig:antmaze_visualization}d).

\begin{figure}[t]
    \centering
    \includegraphics[width=0.6\columnwidth]{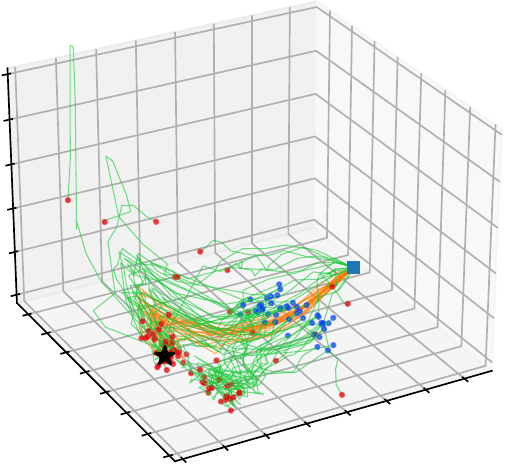}
    \caption{The visualization of the Sawyer Reaching dataset shows blue dots for start positions and red dots for end positions of trajectories. Orange lines indicate ``completed'' trajectories, while green lines represent ``play'' trajectories. A blue square marks the initial position, and a black star marks the target region during evaluation.}
    \label{fig:sawyer_dataset}
    \vskip -0.2in
\end{figure}

\subsection{Sawyer Reaching Task}
\begin{figure}[t]
    \centering
    \includegraphics[width=0.49\columnwidth]{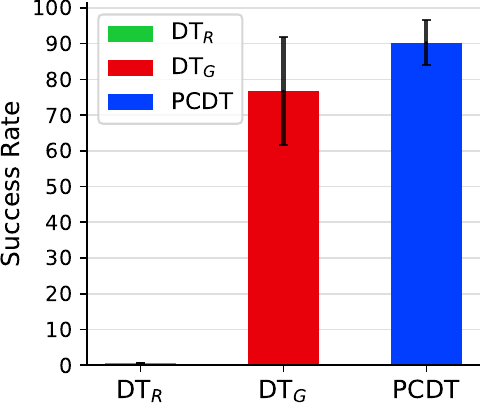}
    \includegraphics[width=0.49\columnwidth]{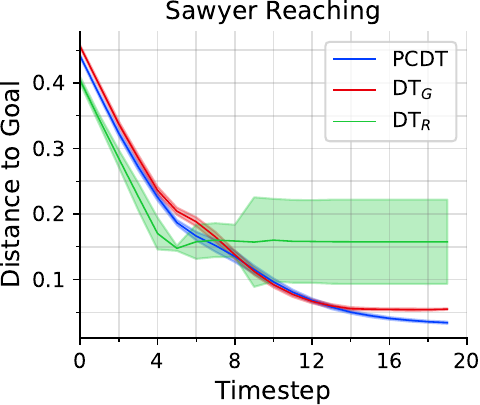}
    \caption{Compasison between DT$_R$, DT$_G$, and \ours \ in terms of the success rate and the distance to the target goal.}
    \label{fig:sawyer_result}
    \vskip -0.2in
\end{figure}

We evaluate the stitching ability of \ours \ on a 7-DoF Sawyer arm (Fig. \ref{fig:environment}) for a goal-reaching task, where the objective is to reach an arbitrary goal. The state space is represented by a 6-D vector, with the first 3 dimensions representing the $x, y, z$ position of the end effector (EE), and the last 3 dimensions representing the $x, y, z$ velocity of the EE. The agent controls the arm by commanding positional translations $(\Delta_x, \Delta_y, \Delta_z)$ of the EE. In each episode, the desired goal is randomly sampled from a specific region, similar to AntMaze. We use a sparse reward function, where $R=1$ when the distance between the current EE's position and the goal is less than $5$cm, and $R=0$ otherwise. The episode's length is set to 20. For the offline dataset, we follow a similar data collection method used in AntMaze Large. Specifically, we collect 110 ``play'' trajectories starting from hand-picked initial positions and reaching specific hand-picked goals, along with 10 ``completed'' trajectories starting from the robot's initial position and ending at the task's goal position, resulting in a total of 120 trajectories. During data collection, a scripted policy is used to generate actions, with random noise added to the initial position, specified goals, and generated actions to encourage diversity. It's worth noting that the goals during collection may differ from those during evaluation. A visualization of a portion of the collected dataset is shown in Fig. \ref{fig:sawyer_dataset}. For this experiment, we use the same hyperparameters as in FrankaKitchen, except we set $k=3$ and $L=8$. We compare the performance of DT$_R$, DT$_G$, and \ours \ agents. The results are presented in Fig. \ref{fig:sawyer_result}. Despite being a simple task, DT$_R$ fails to complete the task when guided by return. In comparison, our method achieves a higher success rate and a smaller final distance to the goal than DT$_G$. This indicates that \ours \ has the potential to enhance performance in tasks involving physical robots compared to DTs, particularly in dealing with unstructured and suboptimal datasets commonly encountered in real-world scenarios.

\section{CONCLUSIONS}

In this work, we introduce Predictive Coding for Decision Transformer (\ours), a method for offline goal-conditioned RL through supervised sequence modeling. By encoding future states into predictive codings, our transformer-based policy can reason future behaviors to reach desired targets. Furthermore, our framework removes the need for rewards from the dataset, enabling training on potentially large unlabeled datasets. Empirical results demonstrate the effectiveness of \ours \ compared to various competitive baselines.

\bibliographystyle{plain} 
\bibliography{IEEEfull,references}

\addtolength{\textheight}{-12cm}   


\end{document}